\documentclass[10pt,twocolumn,letterpaper]{article}

\usepackage{cvpr}
\usepackage{times}
\usepackage{epsfig}
\usepackage{graphicx}
\usepackage{amsmath}
\usepackage{amssymb}
\usepackage{cases}
\usepackage{setspace}
\usepackage[sort,numbers]{natbib}
\usepackage{subeqnarray}
\usepackage{multirow}
\usepackage{tabularx}
\usepackage{subfigure}
\usepackage[pagebackref=true,breaklinks=true,letterpaper=true,colorlinks,bookmarks=false]{hyperref}

\cvprfinalcopy % *** Uncomment this line for the final submission

\begin{document}

%%%%%%%%% TITLE
\title{Learning Deep CNN Denoiser Prior for Image Restoration}

\author{Kai Zhang$^{1,2}$,  Wangmeng Zuo$^{1}$,  Shuhang Gu$^2$,  Lei Zhang$^2$\\
$^1$School of Computer Science and Technology, Harbin Institute of Technology, Harbin, China\\
$^2$Dept. of Computing, The Hong Kong Polytechnic University, Hong Kong, China\\
{\tt\small cskaizhang@gmail.com, wmzuo@hit.edu.cn, shuhanggu@gmail.com, cslzhang@comp.polyu.edu.hk}
}

\maketitle
%\thispagestyle{empty}

%%%%%%%%% ABSTRACT
\begin{abstract}

Model-based optimization methods and discriminative learning methods have been the two dominant strategies for solving various inverse problems in low-level vision.
Typically, those two kinds of methods have their respective merits and drawbacks, e.g., model-based optimization methods are flexible for handling different inverse problems but are usually time-consuming with sophisticated priors for the purpose of good performance; in the meanwhile, discriminative learning methods have fast testing speed but their application range is greatly restricted by the specialized task.
Recent works have revealed that, with the aid of variable splitting techniques, denoiser prior can be plugged in as a modular part of model-based optimization methods to solve other inverse problems (e.g., deblurring). Such an integration induces considerable advantage when the denoiser is obtained via discriminative learning. However, the study of integration with fast discriminative denoiser prior is still lacking. To this end, this paper aims to train a set of fast and effective CNN (convolutional neural network) denoisers and integrate them into model-based optimization method to solve other inverse problems. Experimental results demonstrate that the learned set of denoisers not only achieve promising Gaussian denoising results but also can be used as prior to deliver good performance for various low-level vision applications.

\end{abstract}

%%%%%%%%% BODY TEXT
\section{Introduction}

Image restoration (IR) has been a long-standing problem for its highly practical value in various low-level vision applications~\cite{richardson1972bayesian,andrews1977digital,campisi2016blind}.
In general, the purpose of image restoration is to recover the latent clean image $\mathbf{x}$ from its degraded observation $\mathbf{y} = \mathbf{H}\mathbf{x} + \mathbf{v}$, where $\mathbf{H}$ is a degradation matrix, $\mathbf{v}$ is additive white Gaussian noise of standard deviation $\sigma$.
By specifying different degradation matrices, one can correspondingly get different IR tasks. Three classical IR tasks would be image denoising when $\mathbf{H}$ is an identity matrix, image deblurring when $\mathbf{H}$ is a blurring operator, image super-resolution when $\mathbf{H}$ is a composite operator of blurring and down-sampling.

Since IR is an ill-posed inverse problem, the prior which is also called regularization needs to be adopted to constraint the solution space~\cite{roth2009fields,zoran2011learning}.
From a Bayesian perspective, the solution $\hat{\mathbf{x}}$ can be obtained by solving a Maximum A Posteriori (MAP) problem,
\begin{equation}\label{eq1}
  \hat{\mathbf{x}} = \mathop{\arg}\mathop{\max}\limits_\mathbf{x} ~ \log p(\mathbf{y}|\mathbf{x}) + \log p(\mathbf{x})
\end{equation}
where $\log p(\mathbf{y}|\mathbf{x})$ represents the log-likelihood of observation $\mathbf{y}$, $\log p(\mathbf{x})$ delivers the prior of $\mathbf{x}$ and is independent of $\mathbf{y}$.
More formally, Eqn.~\eqref{eq1} can be reformulated as
\begin{equation}\label{eq2}
  \hat{\mathbf{x}} = \mathop{\arg}\mathop{\min}_\mathbf{x} ~ \frac{1}{2}\|\mathbf{y} - \mathbf{H}\mathbf{x}\|^2 + \lambda \Phi(\mathbf{x})
\end{equation}
where the solution minimizes an energy function composed of a fidelity term $\frac{1}{2}\|\mathbf{y} - \mathbf{H}\mathbf{x}\|^2$, a regularization term $\Phi(\mathbf{x})$ and a trade-off parameter $\lambda$.
The fidelity term guarantees the solution accords with the degradation process, while the regularization term enforces desired property of the output.

Generally, the methods to solve Eqn.~\eqref{eq2} can be divided into two main categories, \ie, model-based optimization methods and discriminative learning methods.
The model-based optimization methods aim to directly solve Eqn.~\eqref{eq2} with some optimization algorithms which usually involve a time-consuming iterative inference.
On the contrary, discriminative learning methods try to learn the prior parameters $\Theta$ and a compact inference through an optimization of a loss function on a training set containing degraded-clean image pairs~\cite{tappen2007utilizing,barbu2009training,sun2013separable,schmidt2014shrinkage,chen2015trainable}. The objective is generally given by
\begin{equation}\label{eq4}%\vspace{-0.1cm}
  \min_\Theta \ell(\mathbf{\hat{x}}, \mathbf{x}) \quad s.t. \quad \hat{\mathbf{x}} = \mathop{\arg}\mathop{\min}_\mathbf{x} ~ \frac{1}{2}\|\mathbf{y} - \mathbf{H}\mathbf{x}\|^2 + \lambda \Phi(\mathbf{x}; \Theta)
\end{equation}
Because the inference is guided by the MAP estimation, we refer to such methods as MAP inference guided discriminative learning methods.
By replacing the MAP inference with a predefined nonlinear function $\mathbf{\hat{x}} = f(\mathbf{y}, \mathbf{H}; \Theta)$, one can treat the plain discriminative learning methods as general case of Eqn.~\eqref{eq4}. It can be seen that one obvious difference between model-based optimization method and discriminative learning method is that, the former is flexible to handle various IR tasks by specifying degradation matrix $\mathbf{H}$, whereas the later needs to use the training data with certain degradation matrices to learn the model.
As a consequence, different from model-based optimization methods which have flexibility to handle different IR tasks, discriminative learning methods are usually restricted by specialized tasks.
For example, model-based optimization methods such as NCSR~\cite{dong2013nonlocally} are flexible to handle denoising, super-resolution and deblurring, whereas discriminative learning methods MLP~\cite{burger2012image}, SRCNN~\cite{dong2016}, DCNN~\cite{xu2014deep} are designed for those three tasks, respectively. Even for a specific task such as denoising, model-based optimization methods (\eg, BM3D~\cite{dabov2007image} and WNNM~\cite{gu2014weighted}) can handle different noise levels, whereas discriminative learning method of~\cite{jain2009natural} separately train a different model for each level.

With the sacrifice of flexibility, however, discriminative learning methods can not only enjoy a fast testing speed but also tend to deliver promising performance due to the joint optimization and  end-to-end training. On the contrary, model-based optimization methods are usually time-consuming with sophisticated priors for the purpose of good performance~\cite{gao2012well}.
As a result, those two kinds of methods have their respective merits and drawbacks, and thus it would be attractive to investigate their integration which leverages their respective merits.
Fortunately, with the aid of variable splitting techniques, such as alternating direction method of multipliers (ADMM) method~\cite{boyd2011distributed} and half-quadratic splitting (HQS) method~\cite{geman1995nonlinear}, it is possible to deal with fidelity term and regularization term separately~\cite{parikh2014proximal}, and particularly, the regularization term only corresponds to a denoising subproblem~\cite{danielyan2010image,heide2014flexisp,venkatakrishnan2013plug}. Consequently, this enables an integration of any discriminative denoisers into model-based optimization methods. However, to the best of our knowledge, the study of integration with discriminative denoiser is still lacking.

This paper aims to train a set of fast and effective discriminative denoisers and integrate them into model-based optimization methods to solve other inverse problems.
Rather than learning MAP inference guided discriminative models, we instead adopt plain convolutional neural networks (CNN) to learn the denoisers, so as to take advantage of recent progress in CNN as well as the merit of GPU computation. Particularly, several CNN techniques, including Rectifier Linear Units (ReLU)~\cite{krizhevsky2012imagenet}, batch normalization~\cite{ioffe2015batch}, Adam~\cite{kingma2014adam}, dilated convolution~\cite{yu2015multi} are adopted into the network design or training.
As well as providing good performance for image denoising, the learned set of denoisers are plugged in a model-based optimization method to tackle various inverse problems.

The contribution of this work is summarized as follows:
\begin{itemize}
  \item We trained a set of fast and effective CNN denoisers. With variable splitting technique, the powerful denoisers can bring strong image prior into model-based optimization methods.
  \item The learned set of CNN denoisers are plugged in as a modular part of model-based optimization methods to tackle other inverse problems. Extensive experiments on
classical IR problems, including deblurring and super-resolution, have demonstrated the merits of integrating flexible model-based optimization methods and fast CNN-based discriminative learning methods.
\end{itemize}

%-------------------------------------------------------------------------
\section{Background}

\subsection{Image Restoration with Denoiser Prior}
There have been several attempts to incorporate denoiser prior into model-based optimization methods to tackle with other inverse problems.
In~\cite{danielyan2012bm3d}, the authors used Nash equilibrium to derive an iterative decoupled deblurring BM3D (IDDBM3D) method for image debluring.   In~\cite{egiazarian2015single}, a similar method which is equipped with CBM3D denoiser prior was proposed for single image super-resolution (SISR). By iteratively updating a back-projection step and a CBM3D denoising step, the method has an encouraging performance for its PSNR improvement over SRCNN~\cite{dong2016}.
In~\cite{danielyan2010image}, the augmented Lagrangian method was adopted to fuse the BM3D denoiser into an image deblurring scheme. With a similar iterative scheme to~\cite{danielyan2012bm3d},
a plug-and-play priors framework based on ADMM method was proposed in~\cite{venkatakrishnan2013plug}.
Here we note that, prior to~\cite{venkatakrishnan2013plug}, a similar idea of plug-and-play is also mentioned in~\cite{zoran2011learning} where a half quadratic splitting (HQS) method was proposed for image denoising, deblurring and inpainting.
In~\cite{heide2014flexisp}, the authors used an alternative to ADMM and HQS, \ie, the primal-dual algorithm~\cite{chambolle2011first}, to decouple fidelity term and regularization term.
Some of the other related work can be found in~\cite{sreehari2015plug,rond2016poisson,brifman2016turning,chan2016plug,teodoro2016image,romano2016little}.
All the above methods have shown that the decouple of the fidelity term and regularization term can enable a wide variety of existing denoising models to solve different image restoration tasks.

We can see that the denoiser prior can be plugged in an iterative scheme via various ways. The common idea behind those ways is to decouple the fidelity term and regularization term.
For this reason, their iterative schemes generally involve a fidelity term related subproblem and a denoising subproblem.
In the next subsection, we will use HQS method as an example due to its simplicity.
It should be noted that although the HQS can be viewed as a general way to handle different image restoration tasks,
one can also incorporate the denoiser prior into other convenient and proper optimization methods for a specific application.

\subsection{Half Quadratic Splitting (HQS) Method}
Basically, to plug the denoiser prior into the optimization procedure of Eqn.~\eqref{eq2}, the variable splitting technique is usually adopted to
decouple the fidelity term and regularization term. In half quadratic splitting method, by introducing an auxiliary variable $\mathbf{z}$, Eqn.~\eqref{eq2} can be reformulated as a constrained optimization problem which is given by
\begin{equation}\label{eq21}
  \hat{\mathbf{x}} = \mathop{\arg}\mathop{\min}_\mathbf{x} ~ \frac{1}{2}\|\mathbf{y} - \mathbf{H}\mathbf{x}\|^2 + \lambda \Phi(\mathbf{z})  \quad s.t. \quad  \mathbf{z} = \mathbf{x}
\end{equation}
Then, HQS method tries to minimize the following cost function
\begin{equation}\label{eq22}
  \mathcal{L}_\mu(\mathbf{x},\mathbf{z}) = \frac{1}{2}\|\mathbf{y} - \mathbf{H}\mathbf{x}\|^2 + \lambda \Phi(\mathbf{z}) +  \frac{\mu}{2}\|\mathbf{z}-\mathbf{x}\|^2
\end{equation}
where $\mu$ is a penalty parameter which varies iteratively in a non-descending order.
Eqn.~\eqref{eq22} can be solved via the following iterative scheme,
\begin{subequations}\label{eq33}
\begin{numcases}{}
\mathbf{x}_{k+1}=\mathop{\arg}\mathop{\min}_{\mathbf{x}} \|\mathbf{y} - \mathbf{H}\mathbf{x}\|^2 + \mu\|\mathbf{x}-\mathbf{z}_k \|^2 \label{eq33_1}\\
\mathbf{z}_{k+1}=\mathop{\arg}\mathop{\min}_{\mathbf{z}} \frac{\mu}{2}\|\mathbf{z}-\mathbf{x}_{k+1}\|^2  + \lambda \Phi(\mathbf{z}) \label{eq33_2}
\end{numcases}
\end{subequations}
As one can see, the fidelity term and regularization term are decoupled into two individual subproblems. Specifically, the fidelity term is associated with a quadratic regularized least-squares problem (\ie, Eqn.~\eqref{eq33_1}) which has various fast solutions for different degradation matrices.
A direct solution is given by
\begin{equation}\label{eqforward}
  \mathbf{x}_{k+1} = (\mathbf{H}^T\mathbf{H}+\mu \mathbf{I})^{-1}(\mathbf{H}^T\mathbf{y}+\mu\mathbf{z}_k)
\end{equation}
The regularization term is involved in Eqn.~\eqref{eq33_2} which can be rewritten as
\begin{equation}\label{eq28}
  \mathbf{z}_{k+1} = \mathop{\arg}\mathop{\min}_{\mathbf{z}} \frac{1}{2(\sqrt{\lambda/\mu})^2}\|\mathbf{x}_{k+1}-\mathbf{z} \|^2  +  \Phi(\mathbf{z})
\end{equation}
According to Bayesian probability, Eqn.~\eqref{eq28} corresponds to denoising the image $\mathbf{x}_{k+1}$ by a Gaussian denoiser with noise level $\sqrt{\lambda/\mu}$.
As a consequence, any Gaussian denoisers can be acted as a modular part to solve Eqn.~\eqref{eq2}.
To address this, we rewrite Eqn.~\eqref{eq28} by following
\begin{equation}\label{eq29}
  \mathbf{z}_{k+1} = Denoiser(\mathbf{x}_{k+1}, \sqrt{\lambda/\mu})
\end{equation}

It is worth noting that, according to Eqns.~\eqref{eq28} and~\eqref{eq29}, the image prior $\Phi(\cdot)$ can be implicitly replaced by a denoiser prior.
Such a promising property actually offers several advantages. First, it enables to use any gray or color denoisers to solve a variety of inverse problems. Second, the explicit image prior $\Phi(\cdot)$ can be unknown in solving Eqn.~\eqref{eq2}. Third, several complementary denoisers which exploit different image priors can be jointly utilized to solve one specific problem.
Note that this property can be also employed in other optimization methods (\eg, iterative shrinkage/thresholding algorithms ISTA~\cite{bioucas2007new,combettes2005signal} and FISTA~\cite{beck2009fast}) as long as there involves a denoising subproblem.

\section{Learning Deep CNN Denoiser Prior}

\begin{figure*}[htbp]
  \centering
  \includegraphics[width=0.88\textwidth]{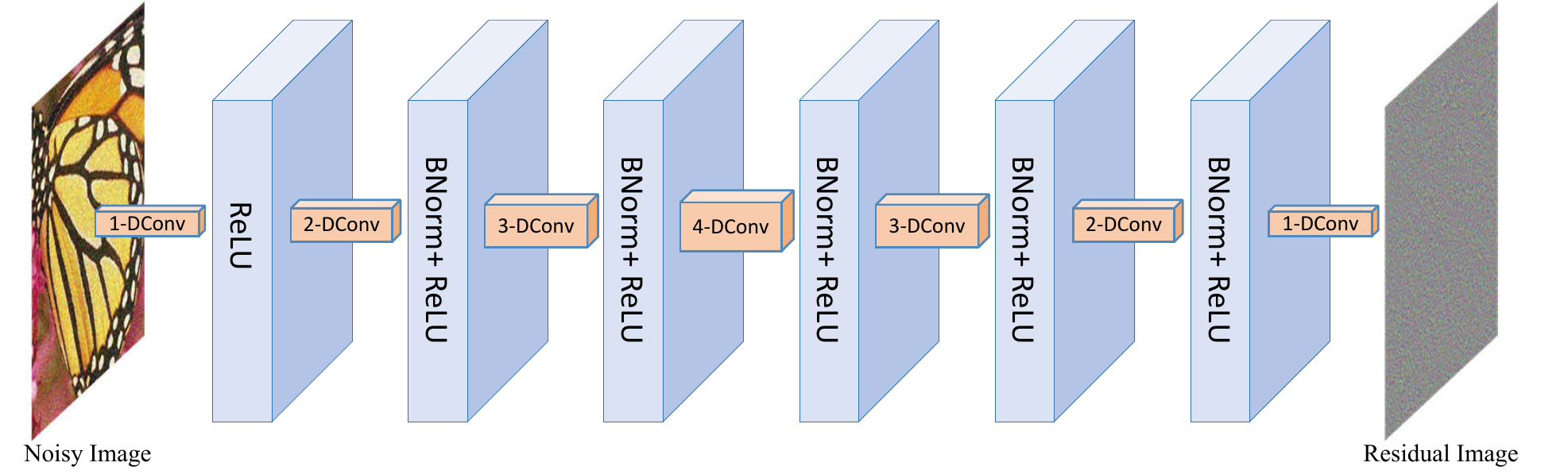}\\
  \caption{The architecture of the proposed denoiser network. Note that ``$s$-DConv'' denotes $s$-dilated convolution~\cite{yu2015multi}, here $s$ = 1, 2, 3 and 4; ``BNorm'' represents batch normalization~\cite{ioffe2015batch}; ``ReLU'' is the rectified linear units ($\max(\cdot, 0)$).}\label{network}
\end{figure*}

\subsection{Why Choose CNN Denoiser?}

As the regularization term of Eqn.~\eqref{eq2} plays a vital role in restoration performance, the choice of denoiser priors thus would be pretty important in Eqn.~\eqref{eq29}. Existing denoiser priors that have been adopted in model-based optimization methods to solve other inverse problems include total variation (TV)~\cite{chambolle2004algorithm,osher2005iterative}, Gaussian mixture models (GMM)~\cite{zoran2011learning}, K-SVD~\cite{elad2006image}, non-local means~\cite{buades2005non} and BM3D~\cite{dabov2007image}.
Such denoiser priors have their respective drawbacks. For example, TV can create watercolor-like artifacts; K-SVD denoiser prior suffers high computational burden; non-local means and BM3D denoiser priors may over-smooth the irregular structures if the image does not exhibit self-similarity property.
Thus, strong denoiser prior which can be implemented efficiently is highly demanded.

Regardless of the speed and performance, color image prior or denoiser is also a key factor that needs to be taken into account. This is because most of the images acquired by modern cameras or transmitted in internet are in RGB format. Due to the correlation between different color channels, it has been acknowledged that jointly handling the color channels tends to produce better performance than independently dealing with each color channel~\cite{foi2006pointwise}. However, existing methods mainly focus on modeling gray image prior and there are only a few works concentrating on modeling color image prior (see, \eg, ~\cite{dabov2007color,mairal2008sparse,rajwade2013image}). Perhaps the most successful color image prior modeling method is CBM3D~\cite{dabov2007color}.
It first decorrelates the image into a luminance-chrominance color space by a hand-designed linear transform and then applies the gray BM3D method in each transformed color channels.
While CBM3D is promising for color image denoising, it has been pointed out that the resulting transformed luminance-chrominance color channels still remain some correlation~\cite{miyata2015inter} and it is preferable to jointly handle RGB channels.
Consequently, instead of utilizing the hand-designed pipeline, using discriminative learning methods to automatically reveal the underlying color image prior would be a good alternative.

By considering the speed, performance and discriminative color image prior modeling, we choose deep CNN to learn the discriminative denoisers.
The reasons of using CNN are four-fold. First, the inference of CNN is very efficient due to the parallel computation ability of GPU. Second, CNN exhibits powerful prior modeling capacity with deep architecture.
Third, CNN exploits the external prior which is complementary to the internal prior of many existing denoisers such as BM3D. In other words, a combination with BM3D is expected to improve the performance.
Fourth, great progress in training and designing CNN have been made during the past few years and we can take advantage of those progress to facilitate discriminative learning.

\subsection{The Proposed CNN Denoiser}
The architecture of the proposed CNN denoiser is illustrated in Figure~\ref{network}.
It consists of seven layers with three different blocks, \ie, ``Dilated Convolution$+$ReLU'' block in the first layer,
five ``Dilated Convolution$+$Batch Normalization+ReLU'' blocks in the middle layers, and ``Dilated Convolution'' block in the last layer.
The dilation factors of (3$\times$3) dilated convolutions from first layer to the last layer are set to 1, 2, 3, 4, 3, 2 and 1, respectively.
The number of feature maps in each middle layer is set to 64.
In the following, we will give some important details in our network design and training.

\begin{spacing}{1.4}
\end{spacing}
\noindent
\textbf{Using Dilated Filter to Enlarge Receptive Field.}
It has been widely acknowledged that the context information facilitates the reconstruction of the corrupted pixel in image denoising.
In CNN, to capture the context information, it successively enlarges the receptive field through the forward convolution operations.
Generally, there are two basic ways to enlarge the receptive field of CNN, \ie, increasing the filter size and increasing the depth.
However, increasing the filter size would not only introduce more parameters but also increase the computational burden~\cite{simonyan2014very}.
Thus, using 3$\times$3 filter with a large depth is popularized in existing CNN network design~\cite{szegedy2015googlenet,he2015deep,kim2015accurate}.
In this paper, we instead use the recent proposed dilated convolution to make a tradeoff between the size of  receptive filed and network depth.
Dilated convolution is known for its expansion capacity of the receptive field while keeping the merits of traditional 3$\times$3 convolution.
A dilated filter with dilation factor $s$ can be simply interpreted as a sparse filter of size (2$s$$+$1)$\times$(2$s$$+$1) where only 9 entries of fixed positions can be non-zeros.
Hence, the equivalent receptive field of each layer is 3, 5, 7, 9, 7, 5 and 3. Consequently, it can be easily obtained that the receptive filed of the proposed network is 33$\times$33.
If the traditional 3$\times$3 convolution filter is used, the network will either have a receptive filed of size 15$\times$15 with the same network depth (\ie, 7) or have a depth of 16 with the same receptive filed (\ie, 33$\times$33). To show the advantage of our design over the above two cases, we have trained three different models on noise level 25 with same training settings.
It turns out that our designed model can have an average PSNR of 29.15dB on BSD68 dataset~\cite{roth2009fields}, which is much better than 28.94dB of 7 layers network with traditional 3$\times$3 convolution filter and very close to 29.20dB of 16 layers network.

\begin{spacing}{1.4}
\end{spacing}
\noindent
\textbf{Using Batch Normalization and Residual Learning to Accelerate Training.}
While advanced gradient optimization algorithms can accelerate training and improve the performance, the architecture design is also an important factor.
Batch normalization and residual learning which are two of the most influential architecture design techniques have been widely adopted in recent CNN architecture designs.
In particular, it has been pointed out that the combination of batch normalization and residual learning is particularly helpful for Gaussian denoising since they are beneficial to each other. To be specific, it not only enables fast and stable training but also tends to result in better denoising performance~\cite{zhang2016beyond}.
In this paper, such strategy is adopted and we empirically find it also can enable fast transfer from one model to another with different noise level.

\begin{spacing}{1.4}
\end{spacing}
\noindent
\textbf{Using Training Samples with Small Size to Help Avoid Boundary Artifacts.}
Due to the characteristic of convolution, the denoised image of CNN may introduce annoying boundary artifacts without proper handling. There are two common ways to tackle with this, \ie, symmetrical padding and zero padding. We adopt the zero padding strategy and wish the designed CNN has the capacity to model image boundary.
Note that the dilated convolution with dilation factor 4 in the fourth layer pads 4 zeros in the boundaries of each feature map.
We empirically find that using training samples with small size can help avoid boundary artifacts. The main reason lies in the fact that, rather than using training patches of large size, cropping them into small patches can enable CNN to see more boundary information.
For example, by cropping an image patch of size 70$\times$70 into four small non-overlap patches of size 35$\times$35, the boundary information would be largely augmented.
We also have tested the performance by using patches of large size, we empirically find this does not improve the performance. However, if the size of the training patch is smaller than the receptive field, the performance would decrease.

\begin{spacing}{1.4}
\end{spacing}
\noindent
\textbf{Learning Specific Denoiser Model with Small Interval Noise Levels.}
Since the iterative optimization framework requires various denoiser models with different noise levels, a practical issue on how to train the discriminative models thus should be taken into consideration. Various studies have shown that if the exact solutions of subproblems (\ie, Eqn.~\eqref{eq33_1} and Eqn.~\eqref{eq33_2}) are difficult or time-consuming to optimize, then using an inexact but fast subproblem solution may accelerate the convergence~\cite{lin2010augmented,zoran2011learning}.
In this respect, their is no need to learn many discriminative denoiser models for each noise level.
On the other hand, although Eqn.~\eqref{eq29} is a denoiser, it has a different goal from the traditional Gaussian denoising. The goal of traditional Gaussian denoising is to recover the latent clean image, however, the denoiser here just acts its own role regardless of the noise type and noise level of the image to be denoised. Therefore, the ideal discriminative denoiser in Eqn.~\eqref{eq29} should be trained by current noise level. As a result, there is tradeoff to set the number of denoisers.
In this paper, we trained a set of denoisers on noise level range $[0, 50]$ and divided it by a step size of 2 for each model, resulting in a set of 25 denoisers for each gray and color image prior modelling.
Due to the iterative scheme, it turns out the noise level range of $[0, 50]$ is enough to handle various image restoration problems.
Especially noteworthy is the number of the denoisers which is much less than that of learning different models for different degradations.

%------------------------------------------------------------------------
\section{Experiments}

The Matlab source code of the proposed method can be downloaded at~\url{https://github.com/cszn/ircnn}.
\subsection{Image Denoising}

It is widely acknowledged that convolutional neural networks generally benefit from the availability of large training data.
Hence, instead of training on a small dataset consisting of 400 Berkeley segmentation dataset (BSD) images of size 180$\times$180~\cite{chen2015trainable}, we collect a large dataset which includes 400 BSD images, 400 selected images from validation set of ImageNet database~\cite{deng2009imagenet} and 4,744 images of Waterloo Exploration Database~\cite{ma2016gmad}.
We empirically find using large dataset does not improve the PSNR results of BSD68 dataset~\cite{roth2009fields} but can slightly improve the performance of other testing images.
We crop the images into small patches of size 35$\times$35 and select $N$=256$\times$4,000 patches for training. As for the generation of corresponding noisy patches, we achieve this by adding additive Gaussian noise to the clean patches during training.
Since the residual learning strategy is adopted, we use the following loss function,
\begin{equation}\label{eq:loss}
  \ell(\Theta) = \frac{1}{2N}\sum_{i=1}^N\|f(\mathbf{y}_i; \Theta) - (\mathbf{y}_i - \mathbf{x}_i) \|_F^2
\end{equation}
where $\{(\mathbf{y}_i, \mathbf{x}_i)\}_{i=1}^N$ represents $N$ noisy-clean patch pairs.
To optimize the network parameters $\Theta$, the Adam solver~\cite{kingma2014adam} is adopted.
The step size is started from 1$e$$-$3 and then fixed to 1$e$$-$4 when the training error stops decreasing.
The training was terminated if the training error was fixed in five sequential epochs.
For the other hyper-parameters of Adam, we use their default setting.
The mini-batch size is set to 256. Rotation or/and flip based data augmentation is used during mini-batch learning. The denoiser models are trained in Matlab (R2015b) environment with MatConvNet package~\cite{vedaldi2015matconvnet} and an Nvidia Titan X GPU.
To reduce the whole training time, once a model is obtained, we initialize the adjacent denoiser with this model. It takes about three days to train the set of denoiser models.

We compared the proposed denioser with several state-of-the-art denoising methods, including two model-based optimization methods (\ie, BM3D~\cite{dabov2007image} and WNNM~\cite{gu2014weighted}), two discriminative learning methods (\ie, MLP~\cite{burger2012image} and TNRD~\cite{chen2015trainable}).
The gray image denoising results of different methods on BSD68 dataset are shown in Table~\ref{table1}.
It can be seen that WNNM, MLP and TNRD can outperform BM3D by about 0.3dB in PSNR.
However, the proposed CNN denoiser can have a PSNR gain of about 0.2dB over those three methods.
Table~\ref{table2} shows the color image denoising results of benchmark CBM3D and our proposed CNN denoiser, it can be seen that the proposed denoiser consistently outperforms CBM3D by a large margin. Such a promising result can be attributed to the powerful color image prior modeling capacity of CNN.

\begin{table}[!tbp]\footnotesize
\caption{The average PSNR(dB) results of different methods on (gray) BSD68 dataset.}
\center
\begin{tabular}{|p{1.4cm}<{\centering}|p{.86cm}<{\centering}|p{.86cm}<{\centering}|p{.86cm}<{\centering}|p{.86cm}<{\centering}|p{.86cm}<{\centering}|p{.86cm}<{\centering}|}
  \hline
 \scriptsize Methods & \scriptsize BM3D& \scriptsize WNNM&\scriptsize TNRD   & \scriptsize MLP & \scriptsize Proposed \\ \hline
  $\sigma = 15$ & 31.07 &  31.37& 31.42  &  - & 31.63  \\\hline
  $\sigma = 25$ & 28.57 &  28.83  & 28.92 & 28.96   & 29.15\\\hline
  $\sigma = 50$ & 25.62 &  25.87  & 25.97 &  26.03 & 26.19 \\
  \hline
\end{tabular}
\label{table1}
\end{table}
\begin{table}[!tbp]\footnotesize
\caption{The average PSNR(dB) results of CBM3D and proposed CNN denoiser on (color) BSD68 dataset.}
\center
\begin{tabular}{|p{1.4cm}<{\centering}|p{.86cm}<{\centering}|p{.86cm}<{\centering}|p{.86cm}<{\centering}|p{.86cm}<{\centering}|p{.86cm}<{\centering}|p{.86cm}<{\centering}|}
  \hline
  Noise Level & 5&  15& 25  & 35 & 50 \\ \hline
  CBM3D & 40.24  &  33.52    & 30.71 &  28.89 & 27.38  \\\hline
  Proposed & 40.36 &  33.86    & 31.16 & 29.50   & 27.86\\
  \hline
\end{tabular}
\label{table2}
\end{table}

For the run time, we compared with BM3D and TNRD due to their potential value in practical applications.
Since the proposed denoiser and TNRD support parallel computation on GPU, we also give the GPU run time. To make a further comparison with TNRD under similar PSNR performance,
we additionally provide the run time of the proposed denoiser where each middle layer has 24 feature maps.
We use the Nvidia cuDNN-v5 deep learning library to accelerate the GPU computation and the memory transfer time between CPU and GPU is not considered.
Table~\ref{table3} shows the run times of different methods for denoising images of size 256$\times$256, 512$\times$512 and 1024$\times$1024 with noise level 25.
We can see that the proposed denoiser is very competitive in both CPU and GPU implementation. It is worth emphasizing that the proposed denoiser with 24 feature maps of each layer has a comparable PSNR of 28.94dB to TNRD but delivers a faster speed. Such a good compromise between speed and performance over TNRD is properly attributed to the following three reasons. First, the adopted 3$\times$3 convolution and ReLU nonlinearity are simple yet effective and efficient. Second, in contrast to the stage-wise architecture of TNRD which essentially has a bottleneck in each immediate output layer, ours encourages a fluent information flow among different layers, thus having larger model capacity. Third, batch normalization which is beneficial to Gaussian denoising is adopted.
According to the above discussions, we can conclude that the proposed denoiser is a strong competitor against BM3D and TNRD.

\begin{table}[!htbp]\footnotesize
\caption{Run time (in seconds) of different methods on images of size 256$\times$256, 512$\times$512 and 1024$\times$1024 with noise level 25.}
\center
\begin{tabular}{|p{1.1cm}<{\centering}|p{.8cm}<{\centering}|p{.9cm}<{\centering}|p{.9cm}<{\centering}|p{1cm}<{\centering}|p{1cm}<{\centering}|}%p{0.1\textwidth}
  \hline
 \scriptsize Size &\scriptsize Device &\scriptsize BM3D   &\scriptsize TNRD  & $\text{\scriptsize Proposed}_{24}$ & $\text{\scriptsize Proposed}_{64}$ \\ \hline
  \multirow{2}{*}{\scriptsize 256$\times$256}& CPU  & 0.66  & 0.47 & 0.10  &  0.310 \\%\cline{2-6}
                                         &  GPU  & -   & 0.010 & 0.006   & 0.012\\\hline
  \multirow{2}{*}{\scriptsize 512$\times$512}& CPU  & 2.91 &1.33   & 0.39  & 1.24 \\%\cline{2-6}
                                         &  GPU  & -   &  0.032 & 0.016   & 0.038\\\hline
  \multirow{2}{*}{\scriptsize 1024$\times$1024}& CPU  & 11.89 & 4.61 &  1.60 & 4.65 \\%\cline{2-6}
                                         &  GPU  & -   & 0.116 &  0.059  & 0.146\\
  \hline
\end{tabular}
\label{table3}
\end{table}

\begin{figure}[!b]
\scriptsize{
\begin{center}
\subfigure[]
{\includegraphics[width=0.076\textwidth]{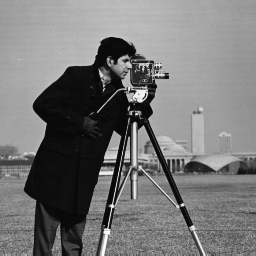}}
\subfigure[]
{\includegraphics[width=0.076\textwidth]{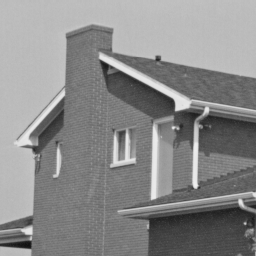}}
\subfigure[]
{\includegraphics[width=0.076\textwidth]{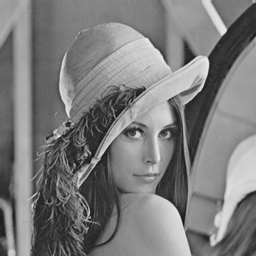}}
\subfigure[]
{\includegraphics[width=0.076\textwidth]{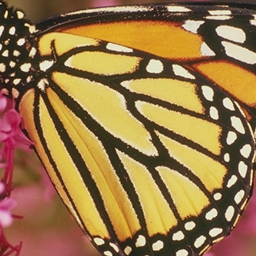}}
\subfigure[]
{\includegraphics[width=0.076\textwidth]{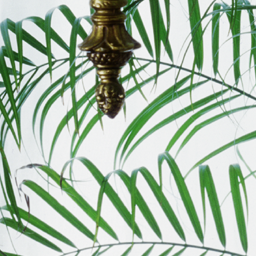}}
\subfigure[]
{\includegraphics[width=0.076\textwidth]{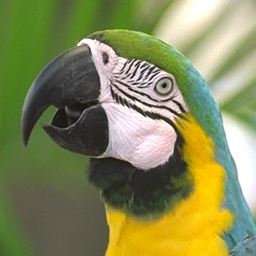}}
\caption{\small Six testing images for image deblurring. (a) \emph{Cameraman}; (b) \emph{House}; (c) \emph{Lena}; (d) \emph{Monarch}; (e) \emph{Leaves}; (f) \emph{Parrots}.}\label{figs6}
\end{center}}\vspace{-.3cm}
\end{figure}

\subsection{Image Deblurring}

\begin{figure*}[!htbp]
\scriptsize{
\begin{center}
\subfigure[Blurry and noisy image]
{\includegraphics[width=0.195\textwidth]{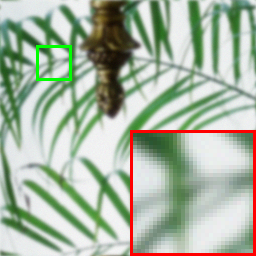}}
\subfigure[IDDBM3D (26.95dB)]
{\includegraphics[width=0.195\textwidth]{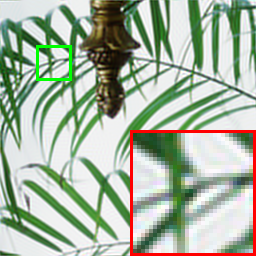}}
\subfigure[NCSR (27.50dB)]
{\includegraphics[width=0.195\textwidth]{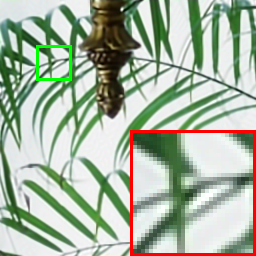}}
\subfigure[MLP (28.91dB)]
{\includegraphics[width=0.195\textwidth]{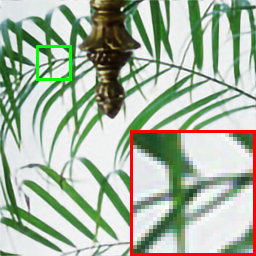}}
\subfigure[Proposed (29.78dB)]
{\includegraphics[width=0.195\textwidth]{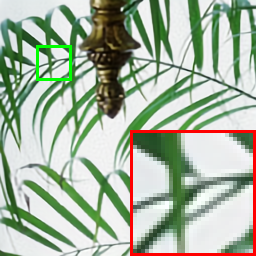}}
\caption{\small  Image deblurring performance comparison for \emph{Leaves} image (the blur kernel is Gaussian kernel with standard deviation 1.6, the noise level $\sigma$ is 2).}\label{fig_blur}
\end{center}}\vspace{-.3cm}
\end{figure*}

As a common setting, the blurry images are synthesized by first applying a blur kernel and then adding additive Gaussian
noise with noise level $\sigma$. In addition, we assume the convolution is carried out with circular boundary conditions.
Thus, an efficient implementation of Eqn.~\eqref{eqforward} by using Fast Fourier Transform (FFT) can be employed.
To make a thorough evaluation, we consider three blur kernels, including a commonly-used Gaussian kernel with standard deviation 1.6 and the first two of the eight real blur kernels from~\cite{levin2009understanding}. As shown in Table~\ref{tab2}, we also consider Gaussian noise with different noise levels.
For the compared methods, we choose one discriminative method named MLP~\cite{schuler2013machine} and three model based optimization methods, including IDDBM3D~\cite{danielyan2012bm3d}, NCSR~\cite{dong2013nonlocally} and EPLL. Among the testing images, apart from three classical gray images as shown in Figure~\ref{figs6}, three color images are also included such that we can test the performance of learned color denoiser prior. In the meanwhile, we note that the above methods are designed for gray image deblurring. Specially, NCSR tackles the color input by first transforming it into YCbCr space and then conducting the main algorithm in the luminance component. In the following experiments, we simply plug the color denoisers into the HQS framework, whereas we separately handle each color channel for IDDBM3D and MLP. Note that MLP trained a specific model for the Gaussian blur kernel with noise level 2.

Once the denoisers are provided, the subsequent crucial issue would be parameter setting. From Eqns.~\eqref{eq33}, we can note that there involve two parameters, $\lambda$ and $\mu$, to tune.
Generally, for a certain degradation, $\lambda$ is correlated with $\sigma^2$ and keeps fixed during iterations, while $\mu$ controls noise level of denoiser. Since the HQS framework is denoiser-based, we instead set the noise level of denoiser in each iteration to implicitly determine $\mu$. Note that the noise level of denoiser $\sqrt{\lambda/\mu}$ should be set from large to small. In our experimental settings, it is decayed exponentially from 49 to a value in $[1, 15]$ depending on the noise level.
The number of iterations is set to 30 as we find it is large enough to obtain a satisfying performance.

The PSNR results of different methods are shown in Table~\ref{tab2}. As one can see, the proposed CNN denoiser prior based optimization method achieves very promising PSNR results.
Figure~\ref{fig_blur} illustrates deblurred \emph{Leaves} image by different methods. We can see that IDDBM3D, NCSR and MLP tend to smooth the edges and
generate color artifacts. In contrast, the proposed method can recover image sharpness and naturalness.

\begin{table}[!b]\footnotesize
\caption{Deblurring results of different methods.}
\center
\begin{tabular}{|p{1cm}<{\centering}|p{0.4cm}<{\centering}|p{.55cm}<{\centering}|p{0.55cm}<{\centering}|p{0.55cm}<{\centering}||p{0.55cm}<{\centering}|p{0.55cm}<{\centering}|p{0.55cm}<{\centering} |}%p{0.1\textwidth}

\hline
 \scriptsize Methods& $\sigma$ &\scriptsize \emph{C.man} &\scriptsize \emph{House}   & \scriptsize \emph{Lena}  & \scriptsize \emph{Monar.}  &\scriptsize \emph{Leaves}  & \scriptsize \emph{Parrots}   \\ \hline
      \multicolumn{8}{|c|}{Gaussian blur with standard deviation 1.6}    \\   \hline% \cline{1-7}
 \scriptsize IDDBM3D  &  \multirow{4}{*}{2} & 27.08  & 32.41  & 30.28 & 27.02  & 26.95 &30.15\\%\cline{2-6}
 \scriptsize NCSR    &    & 27.99  & 33.38 & 30.99   & 28.32  & 27.50 &30.42\\%\cline{2-6
 \scriptsize MLP   &    & 27.84  & 33.43  & 31.10 & 28.87  &  28.91 & 31.24\\%\cline{2-6}
 \scriptsize Proposed  &  & 28.12  & 33.80 & 31.17 &  30.00 & 29.78 & 32.07 \\%\cline{2-6}
  \hline
   \multicolumn{8}{|c|}{Kernel 1 (19$\times$19)~\cite{levin2009understanding}}    \\   \hline% \cline{1-7}
   \scriptsize EPLL   &  \multirow{2}{*}{2.55}  & 29.43  & 31.48 & 31.68 & 28.75  &  27.34 & 30.89\\%\cline{2-6}
 \scriptsize Proposed  &  & 32.07  & 35.17 & 33.88 &  33.62 & 33.92 & 35.49 \\%\cline{2-6}
 \hline
    \scriptsize EPLL   & \multirow{2}{*}{7.65}    & 25.33  & 28.19  & 27.37 & 22.67  &  21.67 & 26.08\\%\cline{2-6}
 \scriptsize Proposed  &  & 28.11  & 32.03 & 29.51 &  29.20 & 29.07 & 31.63 \\%\cline{2-6}

   \hline
       \multicolumn{8}{|c|}{Kernel 2 (17$\times$17)~\cite{levin2009understanding}}    \\   \hline% \cline{1-7}
   \scriptsize EPLL   &  \multirow{2}{*}{2.55}  & 29.67  & 32.26 & 31.00 & 27.53  &  26.75 & 30.44\\%\cline{2-6}
 \scriptsize Proposed  &  & 31.69  & 35.04 & 33.53 &  33.13 & 33.51 & 35.17 \\%\cline{2-6}
 \hline
    \scriptsize EPLL   & \multirow{2}{*}{7.65}    & 24.85  & 28.08  & 27.03 & 21.60 &  21.09 & 25.77\\%\cline{2-6}
 \scriptsize Proposed  &  & 27.70  & 31.94 & 29.27 &  28.73 & 28.63 & 31.35 \\%\cline{2-6}
  \hline

\end{tabular}
\label{tab2}
\end{table}

\subsection{Single Image Super-Resolution}

In general, the low-resolution (LR) image can be modeled by a blurring and subsequent down-sampling operation on a high-resolution one.
The existing super-resolution models, however, mainly focus on modeling image prior and are trained for specific degradation process.
This makes the learned model deteriorates seriously when the blur kernel adopted in training deviates from the real one~\cite{efrat2013accurate,zhang2015revisiting}.
Instead, our model can handle any blur kernels without re-training.
Thus, in order to thoroughly evaluate the flexibility of the CNN denoiser prior based optimization method as well as the effectiveness of the CNN denoisers, following~\cite{peleg2014statistical}, this paper considers three typical image degradation settings for SISR, \ie, bicubic downsampling (default setting of Matlab function $imresize$) with two scale factors 2 and 3~\cite{cui2014deep,dong2016}
and blurring by Gaussian kernel of size 7$\times$7 with standard deviation 1.6
followed by downsampling with scale factor 3~\cite{dong2013nonlocally,peleg2014statistical}.

Inspired by the method proposed in~\cite{egiazarian2015single} which iteratively updates a back-projection~\cite{irani1993motion} step and a denoising step for SISR,
we use the following back-projection iteration to solve Eqn.~\eqref{eq33_1},
\begin{equation}\label{eqsr}
  \mathbf{x}_{k+1} = \mathbf{x}_{k} - \alpha (\mathbf{y} - \mathbf{x}_{k}\downarrow_{\emph{sf}})\uparrow^{\emph{sf}}_{bicubic}
\end{equation}
where $\downarrow_{\emph{sf}}$ denotes the degradation operator with downscaling factor $\emph{sf}$, $\uparrow_{bicubic}^{\emph{sf}}$ represents bicubic interpolation operator with upscaling factor $\emph{sf}$, and $\alpha$ is the step size.
It is worthy noting that the iterative regularization step of methods such as NCSR and WNNM actually corresponds to solving Eqn.~\eqref{eq33_1}. From this viewpoint, those methods are optimized under HQS framework.
Here, note that only the bicubic downsampling is considered in~\cite{egiazarian2015single}, whereas Eqn.~\eqref{eqsr} is extended to deal with different blur kernels.
To obtain a fast convergence, we repeat Eqn.~\eqref{eqsr} five times before applying the denoising step. The number of main iterations is set to 30, the step size $\alpha$ is fixed to 1.75 and the noise levels of denoiser are decayed exponentially from 12$\times$$\emph{sf}$ to $\emph{sf}$.

\begin{table*}[!htbp]\scriptsize
\caption{\small Average PSNR(dB) results of different methods for single image super-resolution on Set5 and Set14.} %Red color indicates the best performance and blue color indicates the second best performance
\center
\begin{tabular}{|p{0.7cm}<{\centering}|p{0.7cm}<{\centering}|p{0.8cm}<{\centering}|p{0.8cm}<{\centering}|p{0.95cm}<{\centering}|p{0.95cm}<{\centering}|p{0.95cm}<{\centering}|p{0.95cm}<{\centering}|p{0.95cm}<{\centering}|p{0.95cm}<{\centering}|p{0.95cm}<{\centering}|p{0.95cm}<{\centering}|p{0.95cm}<{\centering}|}
  \hline
  % after \\: \hline or \cline{col1-col2} \cline{col3-col4} ...\multirow{4}{2cm}{This is a demo table}
Dataset & Scale & Kernel & Channel & SRCNN & VDSR  & NCSR  & SPMSR & SRBM3D  & $\text{SRBM3D}_G$ & $\text{SRBM3D}_C$ & $\text{\scriptsize Proposed}_{G}$ & $\text{\scriptsize Proposed}_{C}$   \\ \hline

\multirow{6}{*}{Set5} & \multirow{2}{*}{2} & \multirow{2}{*}{Bicubic} & Y   & 36.65	 & 37.56  & - & 36.11 & 37.10  & 36.34 & 36.25 & 37.43 & 37.22  \\
                      &                    &                          & RGB & 34.45 & 35.16  & - & 33.94 &  - & 34.11 & 34.22 & 35.05 & 35.07 \\\cline{2-13}
                      & \multirow{2}{*}{3} & \multirow{2}{*}{Bicubic} & Y   & 32.75 & 33.67  & - & 32.31 & 33.30  & 32.62 & 32.54 & 33.39 & 33.18 \\
                      &                    &                          & RGB & 30.72 & 31.50  & - & 30.32 &  - & 30.57& 30.69 & 31.26 & 31.25 \\\cline{2-13}
                      & \multirow{2}{*}{3} & \multirow{2}{*}{Gaussian}& Y   & 30.42 & 30.54  & 33.02 & 32.27 & - & 32.66 & 32.59 & 33.38 & 33.17 \\
                      &                    &                          & RGB & 28.50 & 28.62  & 30.00 & 30.02 &  - & 30.31 & 30.74 & 30.92  & 31.21\\\hline

\multirow{6}{*}{Set14} & \multirow{2}{*}{2} & \multirow{2}{*}{Bicubic} & Y   & 32.43 & 33.02  & - & 31.96 & 32.80  & 32.09 & 32.25 &32.88 &  32.79 \\
                      &                    &                          & RGB & 30.43 & 30.90  & - & 30.05 & -  &  30.15& 30.32 & 30.79 &  30.78 \\\cline{2-13}
                      & \multirow{2}{*}{3} & \multirow{2}{*}{Bicubic} & Y   & 29.27 & 29.77  & - &28.93  & 29.60  & 29.11 & 29.27 & 29.61 & 29.50 \\
                      &                    &                          & RGB & 27.44 & 27.85 & - & 27.17 &  - & 27.32 & 27.47 & 27.72 & 27.67 \\\cline{2-13}
                      & \multirow{2}{*}{3} & \multirow{2}{*}{Gaussian}& Y   & 27.71 & 27.80  & 29.26 & 28.89 &  - & 29.18 & 29.39 &29.63   & 29.55 \\
                      &                    &                          & RGB & 26.02 & 26.11  & 26.98 & 27.01 &  - & 27.24 &27.60 & 27.59  & 27.70\\\hline
\end{tabular}
\label{table:SR}%\vspace{-.2cm}
\end{table*}

\begin{figure*}[!htbp]
\scriptsize{
\begin{center}
\subfigure[Ground-truth]
{\includegraphics[width=0.195\textwidth]{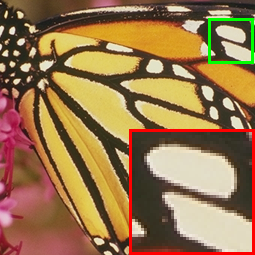}}
\subfigure[Zoomed LR image]
{\includegraphics[width=0.195\textwidth]{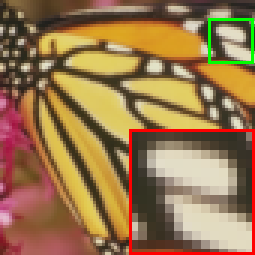}}
\subfigure[SRCNN (24.46dB)]
{\includegraphics[width=0.195\textwidth]{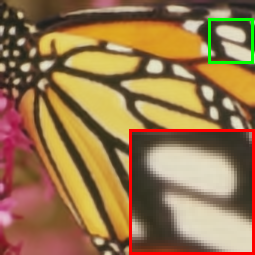}}
\subfigure[VDSR (24.73dB)]
{\includegraphics[width=0.195\textwidth]{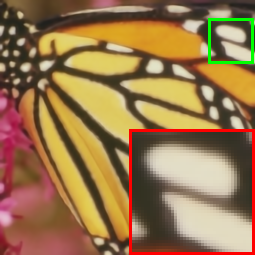}}
\subfigure[$\text{Proposed}_{G}$ (29.32dB)]
{\includegraphics[width=0.195\textwidth]{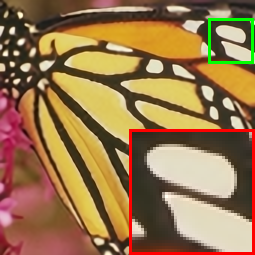}}
\caption{\small  Single image super-resolution performance comparison for \emph{Butterfly} image from Set5 (the blur kernel is 7$\times$7 Gaussian kernel with standard deviation 1.6, the scale factor is 3). Note that the comparison with SRCNN and VDSR is unfair. The proposed deep CNN denoiser prior based optimization method can super-resolve the LR image by tuning the blur kernel and scale factor without training, whereas SRCNN and VDSR need additional training to deal with such cases. As a result, this figure is mainly used to show the flexibility advantage of the proposed deep CNN denoiser prior based optimization method over discriminative learning methods.}\label{fig_sr}
\end{center}}\vspace{-.2cm}
\end{figure*}

The proposed deep CNN denoiser prior based SISR method is compared with five state-of-the-art methods, including two CNN-based discriminative learning methods (\ie, SRCNN~\cite{dong2016} and VDSR~\cite{kim2015accurate}), one statistical prediction model based discriminative learning method~\cite{peleg2014statistical} which we refer to as SPMSR, one model based optimization method (\ie, NCSR~\cite{dong2013nonlocally}) and one denoiser prior based method (\ie, SRBM3D~\cite{egiazarian2015single}).
Except for SRBM3D, all the existing methods conducted their main algorithms on Y channel (\ie, luminance) of transformed YCbCr space. In order to evaluate the proposed color denoiser prior, we also conduct experiments on the original RGB channels and thus the PSNR results of super-resolved RGB images of different methods are also given. Since the source code of SRBM3D is not available, we also compare two methods which replace the proposed CNN denoiser with BM3D/CBM3D denoiser. Those two methods are denoted by $\text{SRBM3D}_G$ and $\text{SRBM3D}_C$, respectively.

Table~\ref{table:SR} shows the average PSNR(dB) results of different methods for SISR on Set5 and Set14~\cite{timofte2014a}.
Note that SRCNN and VDSR are trained with bicubic blur kernel, thus it is unfair to use their models to super-resolve the low-resolution image with Gaussian kernel. As a matter of fact, we give their performances to demonstrate the limitations of such discriminative learning methods.
From Table~\ref{table:SR}, we can have several observations. First, although SRCNN and VDSR achieve promising results to tackle the case with bicubic kernel, their performance deteriorates seriously when the low-resolution image are not generated by bicubic kernel (see Figure~\ref{fig_sr}). On the other hand, with the accurate blur kernel, even NCSR and SPMSR outperform SRCNN and VDSR for Gaussian blur kernel. In contrast, the proposed methods (denoted by $\text{Proposed}_{G}$ and $\text{Proposed}_{C}$) can handle all the cases well. Second, the proposed methods have a better PSNR result than $\text{SRBM3D}_C$ and $\text{SRBM3D}_G$ which indicates good denoiser prior facilitates to solve super-resolution problem.
Third, both of the gray and color CNN denoiser prior based optimization methods can produce promising results.
As an example for the testing speed comparison, our method can super-resolve the \emph{Butterfly} image in 0.5 second on GPU and 12 seconds on CPU, whereas NCSR spends 198 seconds on CPU.

\section{Conclusion}

In this paper, we have designed and trained a set of fast and effective CNN denoisers for image denoising. Specially, with the aid of variable splitting technique, we have plugged the learned denoiser prior into a model-based optimization method of HQS to solve the image deblurring and super-resolution problems. Extensive experimental results have demonstrated that the integration of model-based optimization method and discriminative CNN denoiser results in a flexible, fast and effective framework for various image restoration tasks. On the one hand, different from conventional model-based optimization methods which are usually time-consuming with sophisticated image priors for the purpose of achieving good results, the proposed deep CNN denoiser prior based optimization method can be implemented effectively due to the plug-in of fast CNN denoisers. On the other hand, different from discriminative learning methods which are specialized for certain image restoration tasks, the proposed deep CNN denoiser prior based optimization method is flexible in handling various tasks while can produce very favorable results.
In summary, this work highlights the potential benefits of integrating flexible model-based optimization methods and fast discriminative learning methods.
In addition, this work has shown that learning expressive CNN denoiser prior is a good alternative to model image prior.

While we have demonstrated various merits of plugging powerful CNN denoiser into model-based optimization methods, there also remain room for further study. Some research directions are listed as follows. First, it will be interesting to investigate how to reduce the number of the discriminative CNN denoisers and the number of whole iterations.
Second, extending the proposed CNN denoiser based HQS framework to other inverse problems such as inpainting and blind deblurring would be also interesting.
Third, utilizing multiple priors which are complementary to improve performance is certainly one promising direction.
Finally, and perhaps most interestingly, since the HQS framework can be treated as a MAP inference, this work also provides some insights into designing CNN architecture for task-specific discriminative learning. Meanwhile, one should be aware that CNN has its own design flexibility and the best CNN architecture is not necessarily inspired by MAP inference.

\section{Acknowledgements}
This work is supported by HK RGC General Research Fund (PolyU 5313/13E) and National Natural Science Foundation of China (grant no.~61672446, 61671182).
We gratefully acknowledge the support from NVIDIA Corporation for providing us the Titan X GPU used in this research.

\clearpage

{\small
\bibliographystyle{ieee}
\bibliography{egbib}
}

\end{document}